\documentclass{article}

\usepackage[final,nonatbib]{ap_2018}

\usepackage{comment}
\usepackage[utf8]{inputenc} % allow utf-8 input
\usepackage[T1]{fontenc}    % use 8-bit T1 fonts
\usepackage{hyperref}       % hyperlinks
\usepackage{url}            % simple URL typesetting
\usepackage{booktabs}       % professional-quality tables
\usepackage{amsfonts}       % blackboard math symbols
\usepackage{nicefrac}       % compact symbols for 1/2, etc.
\usepackage{microtype}      % microtypography
\usepackage{xcolor}
\usepackage{graphicx}

\usepackage{caption}
\usepackage{wrapfig}

\title{Tartan: A retrieval-based socialbot powered by a dynamic finite-state machine architecture}

\author{
    \normalsize
    \textbf{George Larionov},
    \textbf{Zachary Kaden},
    \textbf{Hima Varsha Dureddy},\\
    \normalsize
    \textbf{Gabriel Bayomi T. Kalejaiye},
    \textbf{Mihir Kale},
    \textbf{Srividya Pranavi Potharaju},
    \textbf{Ankit Parag Shah}, \\
    \normalsize
    \textbf{Alexander I Rudnicky}\\
    \normalsize
    Language Technologies Institute \\
    Carnegie Mellon University \\
  \texttt{\{glariono, zkaden, hdureddy\}@cs.cmu.edu} \\
  \texttt{\{gbayomi, mihirsak, spothara, aps1\}@cs.cmu.edu} \\
  \texttt{\{air\}@cs.cmu.edu}
}

\makeatletter
\def\thanks#1{\protected@xdef\@thanks{\@thanks
        \protect\footnotetext{#1}}}
\makeatother

\begin{document}

\maketitle

\begin{abstract}
  This paper describes the Tartan conversational agent built for the 2018 Alexa Prize Competition. Tartan is a non-goal-oriented socialbot focused around providing users with an engaging and fluent casual conversation. Tartan's key features include an emphasis on structured conversation based on flexible finite-state models and an approach focused on understanding and using conversational acts. To provide  engaging conversations, Tartan blends script-like yet dynamic responses with data-based generative and retrieval models. Unique to Tartan is that our dialog manager is modeled as a dynamic Finite State Machine. To our knowledge, no other conversational agent implementation has followed this specific structure.
\end{abstract}

\section{Introduction}
Over the past several years, conversational systems have become of increasing interest to the research community. Previously, interest in human-machine spoken language interaction had focused on goal-oriented dialog systems. Over that time, spoken dialog technologies were able to achieve a level of maturity that enabled their use in commercial systems, even as the approaches shifted from knowledge-engineering to machine learning, based in part on the availability of increasingly larger corpora of human-machine interaction. For the latter, see the overview by \cite{YOUNG2010150}.

Goal-oriented interaction, however, represents only one aspect of natural human communication. Left out of sustained investigation were interactions that did not have concrete transactional goals and took place for other reasons, for example social bonding or entertainment. An early example of a socialbot was ELIZA \cite{weizenbaum1966eliza}, a rule-based system that simulated natural conversation.
It's successors include the more recent A.L.I.C.E. \cite{shawar2002comparison} system as well as various systems built for the Loebner Prize competitions \cite{bradevsko2012survey}. In the past few years, a renewed interest in such systems has emerged, making use of readily available data. The initial systems \cite{banchs2012iris} were based on information retrieval techniques and operated by taking immediately preceding inputs (such as the human's last turn, possibly in combination with the system's last turn) as a retrieval key for a database of conversational material, such as film scripts or online chats. While such systems could carry on conversations, they tended to have difficulty maintaining continuity and contextual awareness and would often reduce to question-answering. An alternative approach grew out of the machine learning community, applying deep learning techniques to the problem \cite{1506.05869,sordoni2015neural}. Such systems usually use sequence-to-sequence models, and a variety of approaches have been tried, such as using a hierarchical architecture \cite{serban2015building}. Results have been mixed, often the performance of such systems is greatly hindered by their lack of context and continuity, two very complex concepts not easily learned by a sequence-to-sequence model, possibly due to the inconsistency and relative paucity of data, especially when compared to other fields where machine learning has been successfully applied.

Socialbots in the 2017 Alexa Prize Competition made use of such techniques \cite{DBLP:journals/corr/abs-1801-03604}, but ultimately with limited success. What did work well were approaches based on finite-state machines (FSMs). These would lead the human participant through a more scripted interaction, reminiscent of earlier directed dialog techniques developed for goal-oriented systems. In the most extreme cases, the human would be exclusively answering questions posed by the agent, but more flexible and dynamic FSMs are also possible. That said, a well-scripted static FSM can be quite compelling and can produce a good experience for the user despite not always being characteristic of a natural human conversation, in which the roles are more evenly balanced and each participant is expected to take some initiative in developing the conversation.

In defining our initial approach to the challenge, we acknowledged the need to have FSMs manage parts of the interaction, but we were more interested in creating a framework that could support more balanced conversations. Conversational systems at this point in time need to incorporate scripted interaction grounded in some particular domain (say movies or sports), but they also need to detect and understand conversational signals from the human (e.g. expressions of confusion or praise) and be able to generate their own such signals in appropriate contexts. For example, acknowledgment is key to communicating a sense of engagement (``paying attention'') and allowing the listener to maintain a useful model of their counterpart's internal state. To that end, we took inspiration from early work on conversation, in particular, that of \cite{sacks1974simplest} and \cite{ventola1979structure}. The main observations in these works focused on how interlocutors manage the mechanics of conversation, and on how casual conversation reflects an understanding shared and accepted by both interlocutors. We need to point out that this earlier work is observational and cannot translate directly into a specific computational approach; but it does provide a framework for developing conversational systems and sheds light on the overall structure of human conversation, mimicking which we hold as our ultimate goal.

Tartan is focused around two main dialog strategies: it makes use of FSMs to provide for locally cohesive structure (such as an introduction and topic-specific episodes) and attempts to introduce less constrained responses through the use of a variety of generation and retrieval techniques. Our approach utilizes an array of FSMs which when triggered provide a robust but somewhat constrained conversation and can be swapped in and out based on input from the user. Tartan also makes use of an intent detection module that identifies various conversational markers expressed by the human and guides the conversation in a way which attempts to be reactive on a turn-by-turn basis. Simultaneously, less constrained, and therefore more novel, responses are available to the system via a set of response generators with access to various databases and resources, which can be triggered to provide a completely un-scripted experience.

In this paper, we discuss Tartan's architecture, our approach to natural language processing and response selection, as well as our FSM architecture. We will present some analysis of conversations and summarize our lessons learned.

\section{Related Work}
\begin{figure}[h]
\centering
\includegraphics[width=1.0\textwidth, height=0.2\textwidth]{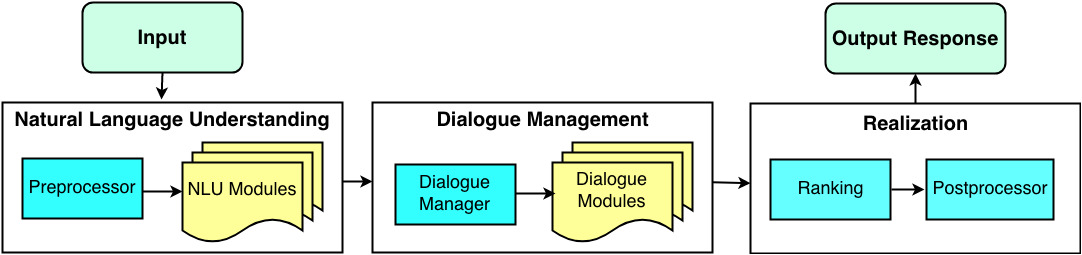}
\caption{A generic socialbot system architecture}
\label{figure:generic_archit}
\end{figure}

This is the second year of the Alexa Prize competition. As such, we have had the fortune to study a previous cohort of socialbots and learn from their strengths and weaknesses. Sounding Board, the inaugural Alexa Prize competition champions, displayed the strength of a robust dialog manager and diverse information retrieval modules \cite{chatbot_wash}.  Alquist, another 2017 Alexa Prize competitor, suggests utilizing information aggregation to store a database of static information that can easily be queried by a socialbot \cite{chatbot_alquist}. They do this by generating knowledge bases of topics, and by continuously expanding their bot's knowledge by introducing new information each day created from news outlets and social media. Many of the 2017 Alexa Prize socialbots were developed with similar architectures (see Figure \ref{figure:generic_archit}).
First, Automated Speech Recognition translates the speech into text. Next, the text is processed by a Natural Language Understanding module. The Natural Language Understanding module contains features such as intent recognition, topic and domain detection, anaphora resolution, and Named Entity Recognition \cite{DBLP:journals/corr/abs-1801-03604}. Some teams augmented this NLU with other features. Sounding Board augments its Natural Language Understanding with an error handling module, which controls for errors in Automated Speech Recognition and Natural Language Processing \cite{chatbot_wash}. Next, the Natural Language Unit hands off information to the Dialog Manager. The Dialog Manager tracks the context and history of the conversation. Slugbot's Dialog Manager modeled dialog flow as a state graph \cite{slugbot}. Magnus built multiple Finite State Machines to model dialog within a specific topic (e.g. movies) \cite{magnus}. We took inspiration from Magnus and built a dynamic Finite State Machine to model dialog flow. Previous Alexa Prize teams primarily used three techniques for response generation: neural text generation, information retrieval, and templates. The teams ensembled multiple response generators. The MILA Team, for example, utilized 22 different response models in their bot \cite{mila}. Ultimately, a response is chosen using a selection strategy and a response output manager. Most teams used a rule-based logic system to determine their response ranking. Meanwhile, the MILA team utilized reinforcement learning for its response ranking \cite{mila}. Some teams, such as Sounding Board, used their response output manager to control prosody \cite{chatbot_wash}.

\section{System Architecture}
\begin{figure}[h]
\centering
\includegraphics[width=0.8\textwidth, height=0.6\textwidth]{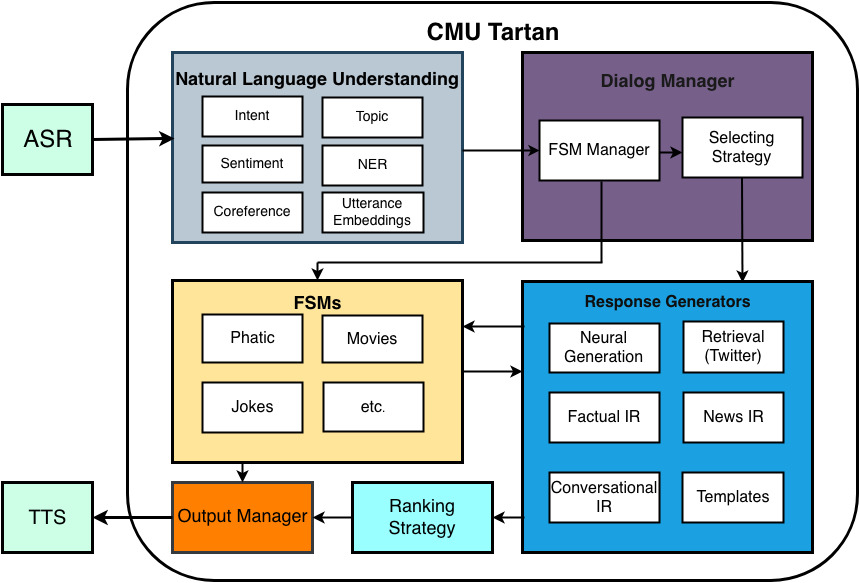}
\caption{System Architecture Diagram}
\label{figure:archit}
\end{figure}

\subsection{Overview}
Figure \ref{figure:archit} shows the design of the Tartan socialbot. It makes use of the infrastructure provided by the Amazon Alexa Prize team (\textsc{cobot}), which was based on their analysis of system architectures developed during the first competition. The bot is intended to be used via an Alexa device, such as an Echo or a Dot. Participating teams did not have any control over which device was used, or over the circumstances of use. For example, we observed a 460 turn conversation at one point; it was not clear to us what was going on. The front end returns a top-n list of utterance hypotheses, as well as individual word timings, plus word-level confidence scores. We used the latter for analysis but not in the running system. Tartan was instrumented using the AWS CloudWatch \footnote{\url{https://aws.amazon.com/cloudwatch}} tool, to provide ongoing performance monitoring.

\subsection{Pre-processing and Natural Language Understanding (NLU)}
A well-functioning socialbot needs to understand what users are saying and react accordingly. The infrastructure processes provided several kinds of information that we made use of when generating and selecting the response. These included:
% \begin{itemize}
% \item Intent
% \item Topic
% \item Sentiment
% \item NER/keywords
% \item Utterance encodings
% \item Anaphora resolution
% \item preheating: changing the prior based on context (e.g. 'I guess' mapped to yes\_intent if we asked a yes/no question)
% \end{itemize}

\textbf{Sentiment:} We use VADER (Valence Aware Dictionary and sEntiment Reasoner) \footnote{\url{https://github.com/cjhutto/vaderSentiment}} for sentiment analysis \cite{gilbert2014vader}. VADER has been designed specifically for social media text, which is closer to human dialogue as compared to the data used for training other open source sentiment classifiers such as NLTK \cite{NLTK}.

\textbf{Topic:} Amazon's topic classification API is used to get a probability distribution over a set of topics for each user utterance.

\textbf{NER:} Recognizing the named entities (e.g. person, organization, etc.) in an utterance is an important step to understanding a user's intent. Tartan uses SpaCy \footnote{\url{https://spacy.io/usage/linguistic-features}} to perform named entity recognition.

\textbf{Anaphora resolution:} To understand the correct interpretation of what a user says, pronouns and other referring expressions must be connected to the right mentions. These mentions could be in the same turn or a previous turn. We use NeuralCoref \footnote{\url{https://github.com/huggingface/neuralcoref}}, which implements the deep learning based coreference resolution system from \cite{clark2016improving}.

\textbf{Intent Detection:} At first, when in the context of an FSM expected responses were predicated on one of the Amazon intents\footnote{\url{https://developer.amazon.com/docs/custom-skills/built-in-intent-library.html}} and sentiment extracted from the user utterance. Under other conditions, keywords were used as query terms sent to our response generators. One of our goals was to move beyond this level of understanding.

In examining our conversation data we found that an overwhelming proportion of user intents were labeled as a \texttt{CatchAllIntent}, meaning that the intent classifier could not come up with a confident label. We concluded that the intents that were well recognized were ones that were based on data from goal-directed systems (e.g. utterances such as \textsc{yes, no, stop}). Similarly, topic labels were limited in scope (understandably, given the training data we assume was available).

Coverage for intents and topics is shown in Table \ref{table:intents}. Only items that accumulate to ~95\% of the total are shown. Note that, except for yes, no, stop, few intent labels are informative. To improve coverage we created intents, such as ``yes\_intent'' that include items apparently not covered in the corresponding intents. Likewise, about 70\% of the topic labels are uninformative. (The prevalence of the Movie\_TV topic labels might be attributed to Tartan's reliance on that topic of conversation.) These data are based on Tartan's July 2018 epoch.

\begin{table}
\centering
\begin{tabular}[t]{|l|c|c|c||c|c|c|c|}
\hline
Intent & count & percent & cumul \%& Topic & count & percent & cumul \%\\
\hline
&&&&&&&\\
CatchAllIntent &110593 &33.6 &33.6 &Phatic &197826 &70.1 &70.1 \\
yes\_intent &69283  &21.1 &54.7  &Movies\_TV &22097  &7.8 &77.9 \\
no\_intent &42553  &12.9 &67.7  &Music &9969  &3.5 &81.5 \\
Launch\...Intent &36357  &11.1 &78.7  &Other &9901  &3.5 &85.0 \\
fsm\_request &29284  &8.9 &87.6  &SciTech &6221  &2.2 &87.2 \\
A\...StopIntent &17894 &5.4 &93.1  &Celebrities &4699  &1.7 &88.9 \\
conclude &5116  &1.6 &94.6  &Sports &4284  &1.5 &90.4 \\
 &&&     &News &3906  &1.4 &91.8 \\
 &&&     &Games &2975  &1.1 &92.8 \\
 &&&     &Pets\_Animals &2929  &1.0 &93.9 \\
 &&&     &Politics &2736  &1.0 &94.8 \\
&&&&&&&\\
\hline
\end{tabular}
\caption{Intents and topics observed over the July 2018 epoch}
\label{table:intents}
\end{table}

\begin{table}
\centering
\begin{tabular}[t]{rl|rl}
\hline
&Gambit&&Favorite\\
\hline\\
145	& SPORTS &       113& COLOR\\
134	& MUSIC&         53 & SONG\\
109	& VIDEO GAMES&   39& BOOK\\
78	& MOVIES&        31&  FOOD\\
64	& BOOKS&         31& ANIMAL\\
50	& IT&            23 & SPORT \\
48	& THE&           18& MOVIES\\
46	& GAMES&         16& SPORTS\\
41	& SEX&           16&ME \\
40	& FOOD&          14& DOGS\\
40	& ANIMALS&       14& ACTOR \\
\\
\hline
\end{tabular}
\caption{Slot values, with observed frequency, found for the concepts  \texttt{Gambit.gambit\_slot} and \texttt{Backstory.backstory\_favorite\_slot}.}
\label{table:slots}
\end{table}

As a consequence, we developed our own set of intents. Practically, we were happy to make use of the existing Amazon intents and our extensions; we therefore concentrated on those utterances that were labeled \texttt{CatchAllIntent}. Specifically, our goal was to identify intents that represented communication about the conversation, as opposed to specific topics of conversation. To accomplish this goal, we represented this information in the form of a semantic grammar. While such grammars have been shown to be useful for goal-directed systems \cite{Ward1994RecentII}, it was not immediately clear the approach would work in the current case. We conjectured that the ``language'' of conversation would, for practical purposes, be limited enough to make reasonable coverage possible. We made one extension to semantic parsing, to allow for dynamic slots. These are different from pre-specified slots; specifically, certain phrases in the grammar, for example, \textit{what is your favorite} will have immediately following material (words not already captured as a different concept) available along with the carrier concept. Doing this allows other parts of the system to deal with the material, ideally in a suitable context. We expect that in the long run it will be possible to create parsers that learn these concepts and identify slot values. At this time, however, we first need to develop an understanding of conversational semantics.

Grammar development was based on sessions collected in the last week of June. The data from intermediate-length (7--12 turns) sessions were examined and utterances of interest were entered into a Phoenix semantic grammar \cite{Ward1994RecentII}. Once the initial grammar was constructed, we use it to process the corpus iteratively and identify utterances that were not covered, allowing the grammar to be augmented. The grammar was further modified during July but primarily to make corrections. However, the set of concepts we initially identified remained the same.

Our final grammar captured intents for 62.7\% of \texttt{CatchAllIntent} utterances, computed over all the July 2018 data. Our intent inventory was focused on conversationally relevant concepts and did not make an effort to explicitly capture domain-specific ones (say for sports or news) or ones we believed should otherwise be ignored at this level of understanding (for example a rambling discourse covering several subjects). We expect that there might be topic-specific concepts that it would be useful to capture in the context of particular FSMs but we chose not to pursue this. Examining the residuals of unparsed utterances, we found that much of it appeared to be driven by specific exchanges (for example references to movies) or uninterpretable. We expect that managing such utterances can be done more effectively in the context of a topic-specific FSM.

Users, on average, produced 1.45 concepts in an utterance. As an example, we observed \texttt{$[$Acknowledgment, Assent, Disclosure, Disclosure.disclosure\_slot$]$}, a reasonable succession of concepts for an utterance such as \textit{really || i agree || my favorite movie is \underline{star wars}} (concept bounds are marked by ``||'', slot value is underlined). One advantage of doing this is that it allows the system to separate conversational acts (the first two) from the core act (a disclosure). The bot might formulate a different response if, say, the \textsc{Assent} was a \textsc{Dissent}. As well, the slot value can be made use of as part of a follow-on response.

We identified a total of 37 major concepts, with an additional 74 sub-concepts that provided specializations deemed useful for flow control and response generation (this includes slot phrases).
For example \textsc{Backstory} includes 17 sub-concepts corresponding to things that users asked about and that could have specific answers (such as favorites). The concepts were organized into nine groupings useful for driving conversation understanding; the six most frequent ones are shown in Table \ref{table:concepts}. Together these six groups account for 97\% of observed concepts. One thing to note is the high frequency of Social concepts. In part, this was due to each session beginning with an exchange of greetings. On the other hand, about 18\% of detected concepts were an Address, calling to Alexa by name.

Note that parsing allows for multiple intents to be identified per utterance. Note also that sub-concepts can be designated as ``\texttt{\_slot}'' or ``\texttt{\_focus}''; meaning that unparsed words following the concept phrase are attached to it as a slot value.

\begin{table}
\centering
\begin{tabular}[t]{cccccc}
\hline
Social & Topic Mgmt & Convo State  & Interest & Rejection & Convo Control\\
28.1\% &24.5\% &22.8\% &11.9\% & 5.0\%& 4.7\% \\
\hline \\
{Address}  & {Gambit} &    {Acknowledgment} & {Backstory} & \texttt{Confusion} &{Conclude}\\
{Social}   & {CoreTopic} &     {Concur}         & {Disclosure} & {Rebuttal} & {Elaboration}\\
{OtherBot} & {Question} &   {Dissent}	    & {Preference} & \texttt{Nasty} & {ExplainResponse}\\
         & {ChatGambit}   & {Assent}             & &{Rejection} & {Assertion}\\
         & {TellMe}   & {Approva}l               & &&{Continuation}\\
         & {ActionGambit}   & {Demur}          && &{Repeat}\\
         & {Curious}    &            &&&\\ [1ex]

\hline
\end{tabular}
\caption{Concepts observed over the July 2018 epoch; concepts in a column are listed in decreasing frequency observed.}
\label{table:concepts}
\end{table}

\subsection{Dialog Manager}
Tartan's unique design required a robust way to choose between response generators and FSMs on the fly and to be able to switch from one to the other while remembering the states of all currently active FSMs in order to potentially return to them in the future. For this purpose, we implemented a dialog manager consisting of two main parts, the FSM Manager and the Selecting and Ranking Strategy. The FSM Manager is in charge of all of Tartan's FSMs, while the Selecting Strategy jumps in if no applicable FSMs are found and chooses a list of appropriate response generators to run. The responses from these response generators are then filtered by the Ranking Strategy to output the best possible response. Note that we do not need to filter the FSM responses since they were created specifically to be valid responses.

\subsubsection{Selecting and Ranking Strategy}
\paragraph{Selecting Strategy} We used an intent map from the list of possible intents recognized by our bot to various response generators. The mapping is a many to many function where one or more intents can map to one or more response generators. We learned these mappings empirically with data collected during the competition. The intents can map to both single-turn response generators as well as to specific FSMs; additionally, FSMs have the ability to override the intent mappings dependent on conversational context. The selecting strategy outputs a list of FSMs or response generators from which we generate candidate responses. % The responses are then filtered by the ranking strategy and a final response is chosen, a process detailed in the next paragraph.

\paragraph{Global Ranking Strategy} Tartan divides candidate response generators into two groups: FSMs and response generators. FSMs are queried first and if an FSM response is generated then the response generators are not run, which decreases latency. It is important to note that FSMs have the ability to query response generators; hence, response generators may be queried as part of an FSM response, which is considered an FSM response and not a response generator response.

\paragraph{FSM Ranking Strategy} The FSM Manager contains a ranking mechanism that chooses the correct FSM and FSM response given a list of potential active FSMs and FSM candidate responses. This is detailed in Section 3.4.1.

\paragraph{Response Generator Ranking Strategy} Our response ranker is a two-tiered ranking system. First, we perform a filtering, which removes undesirable responses. This includes responses that are inappropriate, ungrammatical, irrelevant, excessively negative, too long, etc. To perform this soft filtering we run our candidate responses through several NLU modules. First, we leverage the main NLU pipeline that processes incoming user utterances. We also implemented a rule-based grammar checking module, and a neural relevance module that scores the relevance of a candidate response conditioned on the utterance and the previous turn. Additionally, we remove inappropriate responses and/or certain controversial responses. Whether or not a controversial response is removed depends, in part, on the module that generated the response. For example, our controversiality module is much more sensitive to responses retrieved from social media than to responses retrieved from verified news sources. After filtering and scoring candidate responses, we feed our candidate responses into a rule-based ranking model that selects a candidate response conditioned on the response generator, the utterance/response pair, and the soft filtering module.\par

\subsection{Finite State Machines (FSMs)}
Finite state machines (FSMs) were an integral part of Tartan's conversational strategy. FSMs allowed Tartan to more easily maintain context throughout a conversational arc, and facilitated a more structured conversation that Tartan could logically analyze and continue. One challenge of utilizing FSMs was finding an appropriate conversational balance. At one extreme, the bot can completely control the conversation by asking the interlocutor a series of directed questions, while at the other end of the spectrum the bot can instead allow the interlocutor to dictate the conversational arc, while maintaining a set of predefined responses for conversing on specific topics and a logic that determines how the responses may be combined and structured. Both of these strategies have certain advantages and weaknesses. The more the socialbot directs the conversation, the easier it is to maintain context and to generate appropriate responses at each turn. This permits topical conversations that are more significant and penetrating; however, having the socialbot dictate the arc of the conversation can leave the interlocutor feeling overly constrained and often results in a worse user experience. Alternatively, allowing the interlocutor to dictate the conversation flow allows the user too much freedom, and usually results in the user posing questions that the bot is not capable of correctly parsing and responding to. The ideal socialbot strikes a perfect balance between driving the conversation when necessary and also knowing when to back off and allow the user to take control. This, however, assumes that the bot is perfectly capable of handling any utterance the user throws at it. In reality, any current socialbot must often err on the side of caution by providing a bit more structure to the conversation in order to be better able to anticipate the user's response, at the risk of reducing conversational engagement and novelty.

Tartan's FSM framework involves an FSM Manager which correctly activates and hands off between multiple FSMs. Tartan makes use of a Base FSM that manages the introductory exchange and provides a fall-back when other FSMs exit. The fall-back includes proposing a new topic, minor chitchat, as well as solicitation of topics that the user might want to talk about. Apart from the Base FSM, our design incorporates two categories of FSM: topics and interruptions. Topic FSMs include topics such as Movies, Jokes, and Backstory. Interruptions handle various kinds of interjections a user might voice.

\begin{figure}[h]
\centering
\includegraphics[width=1.0\textwidth]{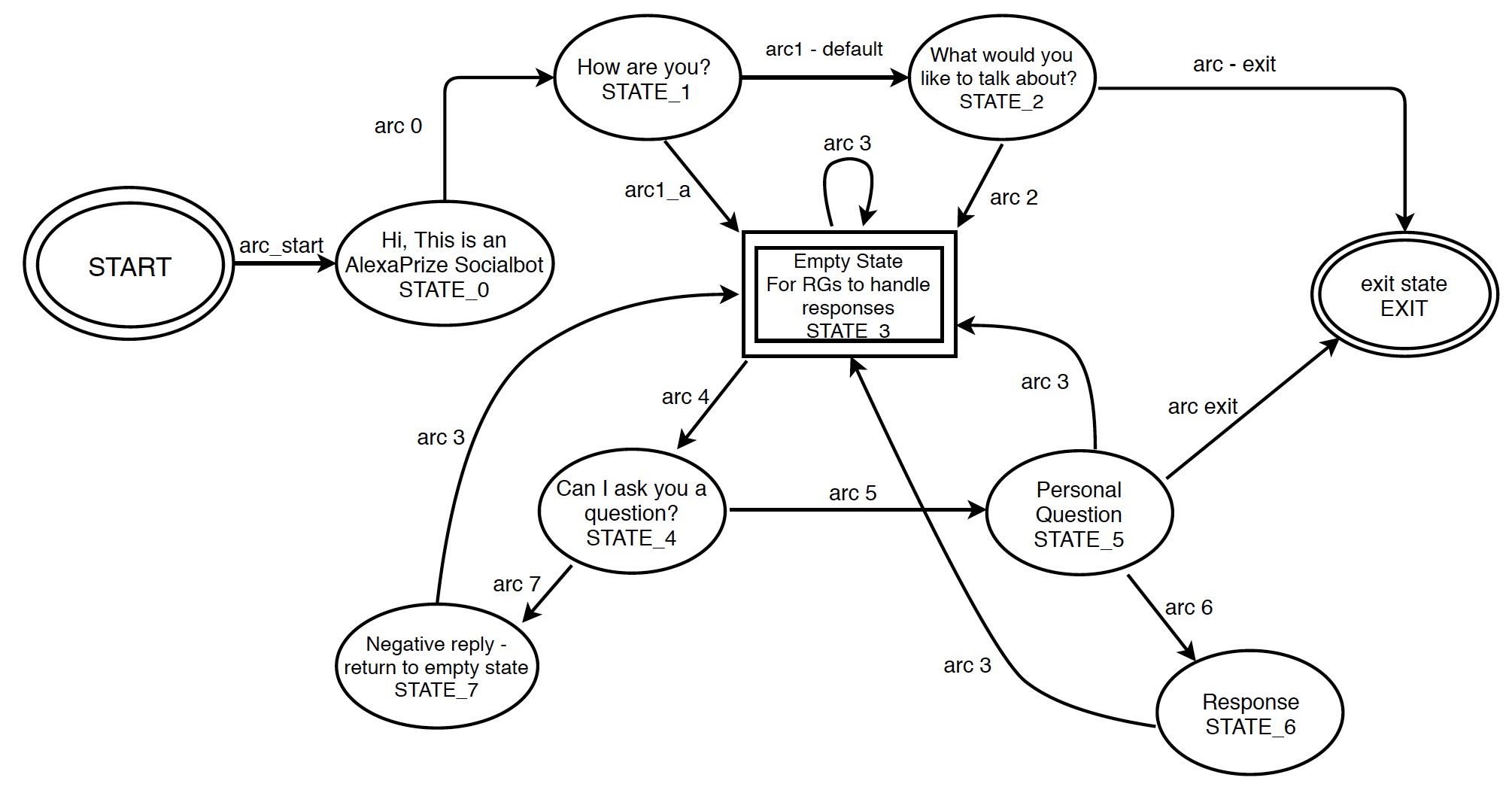}
\caption{Sample diagram of FSM arc structure}
\label{figure:freeform}
\end{figure}

In Figure \ref{figure:freeform} we can see an example arc structure of the Freeform FSM, a successor to a more constrained base FSM (see the Experiments section). The conversation starts in the dummy START state and immediately enters state 0, which greets the user and then follows another null arc (an arc which does not cause a new conversation turn) into state 1, which asks how the user is doing. After the user replies, arc 1 is usually followed, which asks what the user would like to talk about, waits for a response, then enters the "Empty State" which allows different FSMs and response generators to grab the user's utterance and respond to it. Every once in a while, a personal question such as "what is your favorite color?" is attempted, although the user can opt out by declining said question. Not shown here is a state which specifically transitions into another FSM (although the empty state does just that when a user's utterance activates a new FSM). This sort of FSM structure allows for incredibly detailed yet flexible control of a conversation, and we hope that we will soon be able to learn these FSM structures automatically from conversational data gathered throughout this competition.

\subsubsection{FSM Manager}
At the core of Tartan's FSM framework lies the FSM Manager, which utilizes information from the NLU such as intent, sentiment, etc. to choose the correct FSM to respond with and retrieve its response. The FSM Manager first attempts to get a response from the currently active FSM, but failing that it looks for other FSMs which fit the user's utterance. Using the FSM Manager, an FSM can provide a partial response and hand off to a separate FSM to complete it. An FSM can also queue up another FSM or utilize an existing response generator for all or part of its response. Each FSM has a set of functions, called arcs, which coincide with transitions between FSM states. These functions return scores which indicate how well the given arc corresponds to the current context. The score functions take as input the NLU output, the conversation history, and the conversation context and return a real valued score. This allows one or more FSMs to simultaneously propose candidate responses. Finally, the FSM Manager selects a candidate response using a decision tree that reasons over the conversation history, conversation context, active FSM, and FSM candidate response scores.

% Old version below. Changes start in the sentence: 'Each FSM has a set of functions, called arcs, ...'
% At the core of Tartan's FSM framework lies the FSM Manager, which utilizes information from the NLU such as intent, sentiment, etc. to choose the correct FSM to respond with and retrieve its response. The FSM Manager first attempts to get a response from the currently active FSM, but failing that it looks for other FSMs which fit the user's utterance. Using the FSM Manager, an FSM can provide a partial response and hand off to a separate FSM to complete it. An FSM can also queue up another FSM or utilize an existing response generator for all or part of its response. Each FSM has a set of functions, called arcs, which coincide with transitions between FSM states. These functions return scores which indicate how well the given arc corresponds to the current context, which is determined in a rule-based manner using information from the NLU in addition to information about prior turns and/or the current conversation overall. This allows one or more FSMs to simultaneously propose a variety of simultaneous responses, which are then chosen from using a combination of the arc scores and rules concerning which FSMs are allowed to override other FSMs, f.e. an FSM that responds by engaging a user when the user expresses interest in a topic would not override a movie FSM turn even if the user expresses interest in a specific movie, since the combination of arc score of the movie FSM turn and priority of the movie FSM itself override said interruption FSM.

\par
Another key part of the FSM Manager is the FSM stack. This is a data structure which keeps track of which FSM is currently active but most importantly keeps track of which FSMs led to the current one. A significant aspect of the FSM manager is the ability to hand off from one FSM to another, aka the Base FSM can hand off to the movie FSM if the user expresses a desire to talk about movies, and once the Movie FSM exits the Base FSM can be popped off the FSM stack and pick up where it left off. This is important because it models the conversational structure of a human conversation. When people converse they generally instantiate and follow one conversational arc at a time, but they remember the other things that have been brought up in the conversation and are prepared to bring them up when the current arc is completed or interrupted. Of course, our FSM stack cannot fully recreate all the intricacies of a live human conversation, since people often merge arcs and reference world knowledge or perform deductions beyond the capability of any current socialbot, but the FSM stack can maintain conversational context quite well.

\subsubsection{Topic FSMs}
\paragraph{MovieFSM}
The MovieFSM gets triggered when the user's intent is related to movies in general or any specific movie in particular. In order to provide the best possible responses, we compiled a movie database by querying the IMDB database to get the top 500 latest movies, rated by users from around the world and spanning multiple genres such as mystery, horror, comedy, and romance. For each movie we store the title, genre, cast, characters, plot summary, etc. This allows our movie FSM to ask the user questions regarding their favorite characters in the movie, the performance of various acters, and so forth, providing a rich conversational arc with various branches to keep the conversation from getting stale. We also pay attention to how interested the user seems in the current conversation by monitoring utterance-level sentiment and sometimes either proposing a new movie, a new genre, or exiting the movie FSM altogether if the utterances are overwhelmingly negative.

\paragraph{BackStoryFSM}
The Backstory FSM responds to various questions that people want to ask of Tartan (see  Table \ref{table:slots} for some examples). Backstory elements were identified either from sessions or included just for good measure. Based on our examination of early sessions we decided that Tartan should maintain its agent/robot identity and not try to pretend to be too human. For example, in response to questions about family, Tartan answers ``I'm just a bot. I don't have relationships like humans do. Maybe someday...'', but Tartan does have favorite things, like colors and movies. Some responses can have an additional state, e.g. ``and what's your name'' that provides a follow-up.

\paragraph{JokesFSM}
The JokesFSM gets triggered when the NLU indicates the user's intent to hear a joke. To provide an output joke, we needed to first create a database of jokes. This database was created by querying the Internet for top jokes and puns. These queries were deliberately geared towards providing short jokes since we believe during a conversation a user would like to hear a short joke instead of a lengthy story. This decision was informed by reviewing conversation data where often times the user gets noticeably disinterested when the bot produces overly lengthy responses. We collected several dozen of the best jokes and puns that we could find, splitting them up into one-part and two-part jokes. Two-part jokes are jokes which require the user to provide some input (What do you call a lion who never tells the truth? ... "What?" ... The lying king!). The JokesFSM selects a joke at random and outputs it to the user.

\paragraph{SongsFSM}
The SongsFSM provides information about top songs from around the world. Furthermore, it is capable of finding a particular song based on a user request and providing song information to the user. We use the Musixmatch API \footnote{\url{https://www.musixmatch.com}} for song information.

\subsubsection{Interruption FSMs}
The term ``interruption'' is perhaps inaccurate from the perspective of a conversation (it may not act like one) but it does interrupt the currently executing FSM. Interruptions are meant to be short; the user asks a question or makes a comment that should be responded to but should not derail the conversation. The expectation is that under normal circumstances the interrupted FSM (e.g. a topic FSM) will be resumed.
The current set is shown in Table \ref{table:interrup}.

\begin{table}
\centering
\begin{tabular}{|r|l|}
\hline
\textbf{Concur} & ``i know what you mean''  \\
\textbf{Approval} &  ``that would be great''\\
\textbf{Acknowledgment} & ``okay'' ``thanks'' ``i am not surprised'' \\
\textbf{Assent} &``yes of course'' \\
\textbf{Dissent} & ``no i do not like it'' \\
\textbf{Disclosure} & ``i like gardening'' ``today is my birthday''  \\
\textbf{Confusion} & ``you asked me that already'' ``are you there'' \\
\textbf{Praise} & ``you are very smart'' ``that is cute'' \\
\textbf{Boredom} & ``i am bored'' ``you are boring'' \\
\hline
\end{tabular}

\caption{Interruption FSMs in Tartan}
\label{table:interrup}
\end{table}

\subsection{Response Selection and Generation}
We made use of an array of generators to create responses. These include:

\paragraph{Templates}
We identified that many frequently asked questions were non-sequitur; they were usually not asked with the intention of changing or driving the subject of conversation, but were out of curiosity in order to test and explore the socialbot's abilities. Moreover, the interlocutor would often immediately return to discussing the previous subject. We implemented a generalized model utilizing utterance-level embeddings and templated responses. We store pairs of common questions and ideal answers and, given a query, we calculate a similarity metric across all the stored templates. To compute utterance similarity, we calculate the cosine distance of the user utterance with each template. If a similarity metric exceeds our threshold, we respond with the template and return the "ideal" response. In order to build this system, we took advantage of Facebook’s Infersent \cite{infersent} and Google’s Universal Sentence Encoder \cite{google_use} to build our sentence-level representation.

\paragraph{EVI}
EVI is a question answering service provided by Amazon. It provides strong and accurate responses for factual questions. For example, EVI is able to successfully answer questions such as "Who is the president of the United States?" In addition to answering factual questions, EVI was often able to provide superficially strong answers to phatic questions, such as "How is your day going?" While EVI serves its intended purpose quite well, it can struggle when it is deployed in a conversational setting. EVI has no way of setting its context, so it can only reply to the exact text that is passed to it. Additionally, EVI responses are often exact answers to questions posed to it, unlike in natural conversations, where humans often append follow-up questions or content after answering a question in order to facilitate the conversation.

\paragraph{Fact Retrieval}
Fact retrieval presents relevant facts conditioned on past utterances or on the current utterance. In a sense, the fact retrieval response generator can be thought of as a greedy information retrieval module. The motivation behind fact retrieval is to present the user with an interesting fact that minimizes the probability of a user ending the conversation on the following turn. This differs from the rest of our response generators, which are developed to maximize conversation quality across the duration of a conversation. Currently, we are using a domain-specific trivia fact base scraped from IMDB and have focused on movie facts. We periodically output these facts in a conversation when we identify that the user has expressed interest in an entity about which we have a stored fact, or if we identify that a user is particularly disengaged in the conversation. This allows the bot to present the kind of pseudo-random trivia which often comes up in human conversation (e.g. "did you know that ...?"). We plan to expand our database of facts to other domains in order to provide interesting facts and tidbits to the user without repetition.

\paragraph{Neural Generation}
We used a dialog framework developed by FAIR (Facebook AI Research) called ParlAI \cite{parlai}. It utilizes a key-value memory network which essentially memorizes the training data and picks the best response to any given input. The motivation behind using such a network in lieu of a more traditional sequence-to-sequence model was that most sequence-to-sequence models fail to generate grammatically correct or sensible responses most of the time. The keys, in this case, can vary, ranging from entities or keywords to subject-predicate sets of knowledge triples. The network uses efficient memory storage based on hashing which assures efficient and quick retrieval of relevant content via associative search over the memory. This model is, in essence, a general framework for storing and retrieving context in memory based on each scenario. We used a PyTorch implementation of ParlAI trained on PersonaChat - and as a result this model is mainly useful for chit-chat and phatic conversations. This is largely due to the nature of the PersonaChat dataset, which is mostly composed of themed small talk between people adhering to certain assigned personalities. Unfortunately, even this state of the art speech generation approach fails to match a basic FSM in conversational quality due mostly to the lack of context. There has been work in networks which are capable of including context in their generation process, but they are as of yet still not robust enough to be feasible for this particular use case.

\paragraph{Twitter Retrieval}
Twitter is a rich source of information. Tweets are at most 280 characters long and provide brief replies on a diverse range of topics. We used RAKE \cite{RAKE} to extract keywords from the user's utterance and passed these keywords to the Twitter API to retrieve tweets. We then cleaned the tweets, as they often contain hashtags, emoticons and abbreviations. Once ready, the tweets were passed through an extra layer of profanity filtering, as there are many topics which require extra filtering especially from sites like Twitter due to a large amount of profanity and various political arguments.

\paragraph{News Retrieval}
This module focuses on responding to user's questions about current affairs. We extract keywords and named entities from the user's utterance and send these to News API\footnote{\url{https://newsapi.org}}. We focus on headlines in order to respond with popular news for the day sorting the news by popularity. We also get the article description, so that we can elaborate on the headline if prompted.

\paragraph{Conversation Retrieval}
This module aims to respond to phatic utterances such as "How are you?", "What do you want to do today?", etc. To do this, we used the Daily Dialogue dataset \cite{dailydialog}. This dataset contains everyday conversations between two people. We picked question-answer pairs as the training dataset and used StarSpace embeddings \cite{starspace} to encode them. When a user talks to Tartan, we encode the utterance and find the most similar utterance from the conversations and output the response to that utterance. We analyzed last year’s Alexa Prize conversations and found that the best cutoff threshold is 0.82. Initially, we tried encoding all sentences instead of just question-answer pairs. The responses generated were out of context but on-topic. For example, if the query is “I was wondering if I should go hiking today” the response could be “Hiking is my favorite hobby” . The threshold to pick the best response for this was 0.98. The model performed reasonably well during live conversations; however, one possible improvement to this model is to incorporate the user's intent so that the model has more information about the given utterance to work with.

\section{Analysis and Evaluation}
In this section, we discuss interesting results and insights gained throughout our participation in the competition. A number of these findings confirm conventional wisdom about the human conversation. Nevertheless, there has been limited empirical evidence to substantiate that such wisdom extends beyond human-human interactions into human-agent interactions.

One significant result is the impact that an interlocutor's mood has on conversation quality. In the opening turns of a conversation, our socialbot asks the interlocutor how they are doing. We store the interlocutors response and classify their ``mood" along with a spectrum of possible moods. At the extreme ends of this spectrum are ``mood\_unhappy'' and ``mood\_great''. We find that users classified as ``mood\_great'' rate conversations, on average, more than 1.4 points higher than users classified as ``mood\_unhappy.'' This is a striking result given that the rating system ranges from 1-5. This raises an interesting use case for conversational agents. How can conversational agents create positive interactions with interlocutors who are predisposed to be displeased with the conversation? It is likely that an ideal conversation with an unhappy user would be substantially different from an ideal conversation with a happy user. We additionally find that users report being in a great mood roughly 7x more frequently than they report being in an unhappy mood. Given the relative infrequency that users report being unhappy, one must also consider whether this use case is significant enough to warrant major development.

\begin{figure}[h]
\centering
\includegraphics[scale=0.7]{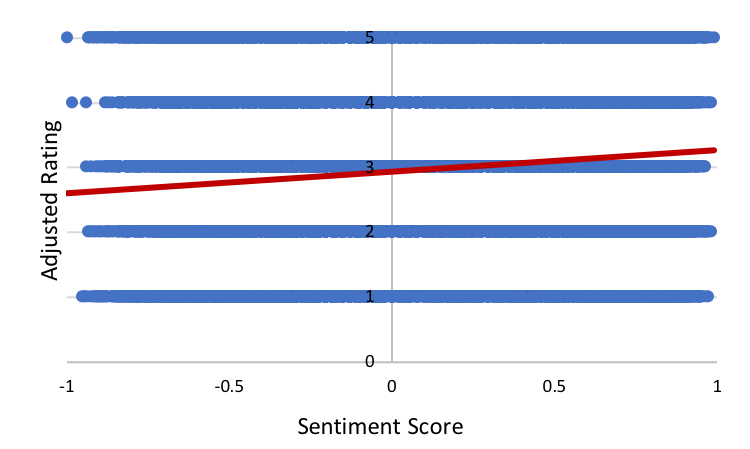}
\caption{A regression analysis of sentiment score against scaled rating. The ratings have been scaled to have a mean value of 3.}
\label{figure:sent_ratings}
\end{figure}

Similarly, we find a number of other metrics that suggest user contentedness correlates with higher evaluations of our agent's conversations. We find that the sentiment of a user utterance correlates with a user's conversation ratings as shown in Figure \ref{figure:sent_ratings}. We rate the sentiment of each user utterance on a scale ranging from -1 to 1, where -1 is most negative, 1 is most positive, and 0 is neutral. We regressed these utterance sentiments against conversation ratings. We find that each increase in sentiment corresponds with a .33 increase in conversation rating. We similarly find that user assents correlate with higher conversation ratings. Conversations with at least one user assent have 9\% higher ratings than conversations with at least one user dissent.

\begin{figure}[h]
\centering
\includegraphics[width=0.7\textwidth, height=0.4\textwidth]{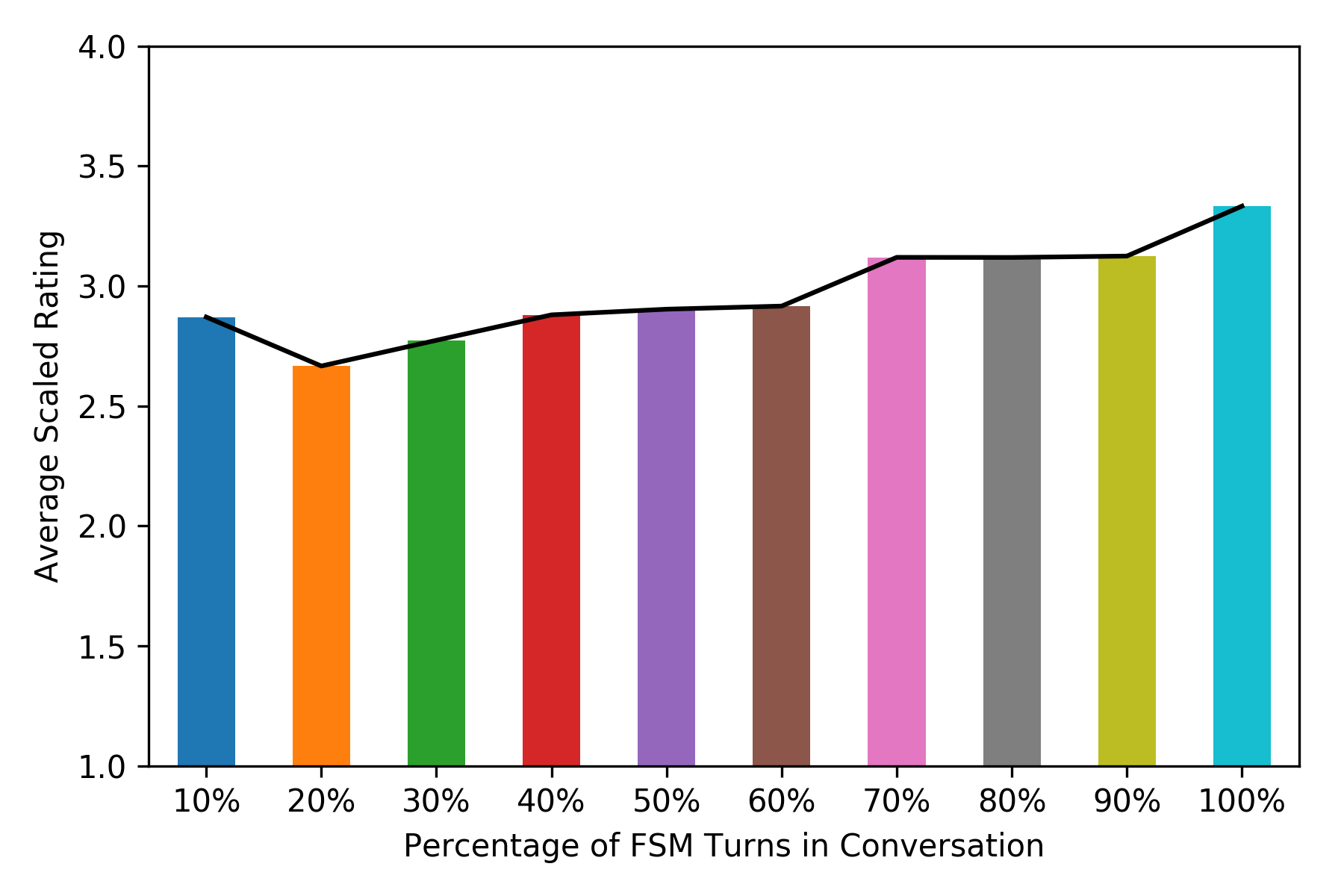}
\caption{Percent of FSM turns in a conversation vs the average scaled conversation rating. The ratings have been scaled to have a mean value of 3.}
\label{figure:percent_fsm_turns}
\end{figure}

Our FSM Manager allows us to examine the average ratings of conversations with respect to what percentage of the turns in each conversation were provided by an FSM. Note that this does not necessarily mean that the output is entirely scripted, some of our FSMs queried databases and outputted different responses based on the input, although the responses were of course more controlled than those provided by one of the standalone response generators. Figure \ref{figure:percent_fsm_turns} shows the relationship between the percentage of a conversation that was FSM-driven and the average rating. For this graph we used conversations of lengths 3 to 15, since we found very short and very long conversations to not be very representative of a user who is genuinely trying to talk to the bot. In the figure, we can see a largely monotonic increase in average conversation rating based on how many FSM turns are contained in a conversation. This is a good sign and supports our theory that while standalone response generators may provide more novel responses than FSMs, overall users prefer the more contextually cohesive conversations provided by the FSMs via the FSM manager.

We also found evidence that user perceptions of conversations are ephemeral. That is, they are heavily dependent on recent conversation history. Alexa Prize competition rules dictated that when an Amazon intent classification detected that a user desired to exit the conversation, the bot should immediately end the conversation without any further discourse. Empirically, we discovered that users would frequently have difficulty exiting a conversation with our bot. To remedy this, we implemented a goodbye module in our socialbot, which, when prompted, would reply with "It sounds like you don't want to talk anymore. Would you like to stop?" After implementing this, conversations with our bot could be ended in two ways: a response could trigger the Amazon classifier and immediately end the conversation, or a response could trigger our socialbot's goodbye script, in which our bot asks the user if he wants to end the conversation and then expresses pleasantries before ending the conversation. We found that conversations that ended with our bot's goodbye script were rated 11\% higher than conversations that terminated via Amazon's classifier. As these two cohorts of conversations were otherwise indistinguishable, it follows that our custom goodbye module is responsible for the increased ratings. Given the simplicity of our goodbye module, we theorize that the content expressed in our goodbye module doesn't improve conversation quality. Rather, we hypothesize that user opinions on conversation quality are mercurial and ephemeral, and, by ending the conversation on a positive tone, we are able to impress upon the interlocutor the appearance of a positive and more natural conversation.

\section{Experiments}
We took the opportunity to explore various potential solutions to specific problems in socialbot operation. This section describes some of these.
\subsection{Controversiality Classifier}

\cite{zhang2018conversations} discusses the important challenges about online social systems and shows that detecting early signs of conversational failure can reduce the prevalence of antisocial behavior, harassment, and personal attacks, which are sadly common on online anonymized forums. In the context of the Alexa Prize, it’s important to flag subtle offensive content to guarantee that the output of our models won’t contribute to whether the conversation will fail and to avoid potential concerning content coming from the user. Therefore, one of the challenges we tackled was the construction of a controversiality classifier. The main idea was to force our bot to take caution whenever the controversiality level of the conversation was higher than usual.

In order to build the classifier, we downloaded 4 years of Reddit data and took advantage of the “controversiality” score given by the API for each user message. Therefore, we could use a supervised approach to learn when to drive the conversation in a different direction given how controversial the topic was. We started by building a Bidirectional LSTM with self-attention classifier, mimicking similar and recent successful approaches for sentiment classification. After tuning and improving our model, we were able to reach the accuracy of 80\% on our validation data; however, our actual results were not consistent with our validation data.

The main reason for the failure of our model was that controversiality varies a lot between different topics. For example, when talking about Nintendo video games on a particular forum, topics about Playstation might be controversial. At first, when we were initially inspecting our data manually we were not able to identify this phenomenon. However, after a few tries we saw that this clear pattern. We decided to filter a huge amount of channels, allowing only the more general ones where topic biases wouldn’t play an important part, in order to reduce the problem. Unfortunately, we were still not able to train an efficient model. For future work, an interesting direction would be to build a topic-sensitive controversiality classifier, which could be even more useful in a socialbot setting.

\subsection{User Embeddings}

To help track context, we experimented with creating personalized user embeddings. These embeddings would track users across their various sessions speaking with Tartan. We hypothesized that we could leverage these user embeddings to make helpful topic suggestions to users. To model this problem, we treated users and user utterances like a traditional sentence-matching problem. In the traditional sentence-matching setting, given a sentence and a set of articles, we wish to find the article that is most likely to contain the sentence. We model our user embeddings such that each user is treated as an article, and each user utterance is treated as a sentence. We trained this model using Facebook AI Research's StarSpace embeddings \cite{starspace} and use the PersonaChat dataset \cite{personachat}, which consists of 164,356 crowdsourced utterances in brief human-human conversations. We found that our model was unable to generalize to live human conversation. On our validation dataset, our model had less than 10\% accuracy when assigning an utterance to a user. Qualitatively, the suggestions provided by leveraging our user embeddings were unsatisfactory and frequently unrelated to the conversation. We theorize that the poor results demonstrated by our model were due to the relatively few utterances spoken by each user in a given conversation. Ultimately, we removed this module and replaced our user embeddings with a bag-of-words model for tracking users. While the bag-of-words model lacked the context of user embeddings, it empirically performed better in the real world.

\subsection{Utterance Embeddings}

We experimented with multiple utterance encodings. We tested Google's Universal Sentence Encoder \cite{google_use}, Allen NLP's ELMo \cite{Peters:2018} on the PersonaChat dataset. Given an utterance, we trained a model to predict the response chosen from the set of all utterances in the dataset. We observed that the PersonaChat dataset is qualitatively different from Alexa conversations because they are artificially generated conversations biased on individual's given persona. Hence, we chose to create our own dataset using more natural conversation corpora. Given an utterance, we generated a candidate list comprising the true response and 12 negative samples from the Daily Dialogue dataset. While generating the dataset, we also made sure not to use candidates which are not exact matches or excessively close to the response. We observed much better results using the embeddings on the custom dataset. The results are displayed in Table \ref{table:embeddings}.

\begin{table}
\centering
\begin{tabular}[t]{|c|c|c|c|c|}
\hline
Dataset & Model & Hits @ 1 & Hits @ 3 & Hits @ 5\\
\hline
PersonaChat & Universal Sentence Encoder & 0.10 & 0.24 & 0.32\\
Custom Dataset & Universal Sentence Encoder & 0.085 & 0.23 & 0.38\\
PersonaChat & ELMo & 0.14 & 0.29 & 0.41\\
Custom Dataset & ELMo & 0.19 & 0.36 & 0.50\\
\hline
\end{tabular}
\caption{Validation results for pretrained utterance embeddings}
\label{table:embeddings}
\end{table}

\subsection{Constrained Base FSM}

\begin{figure}[h]
\centering
\includegraphics[scale=0.6]{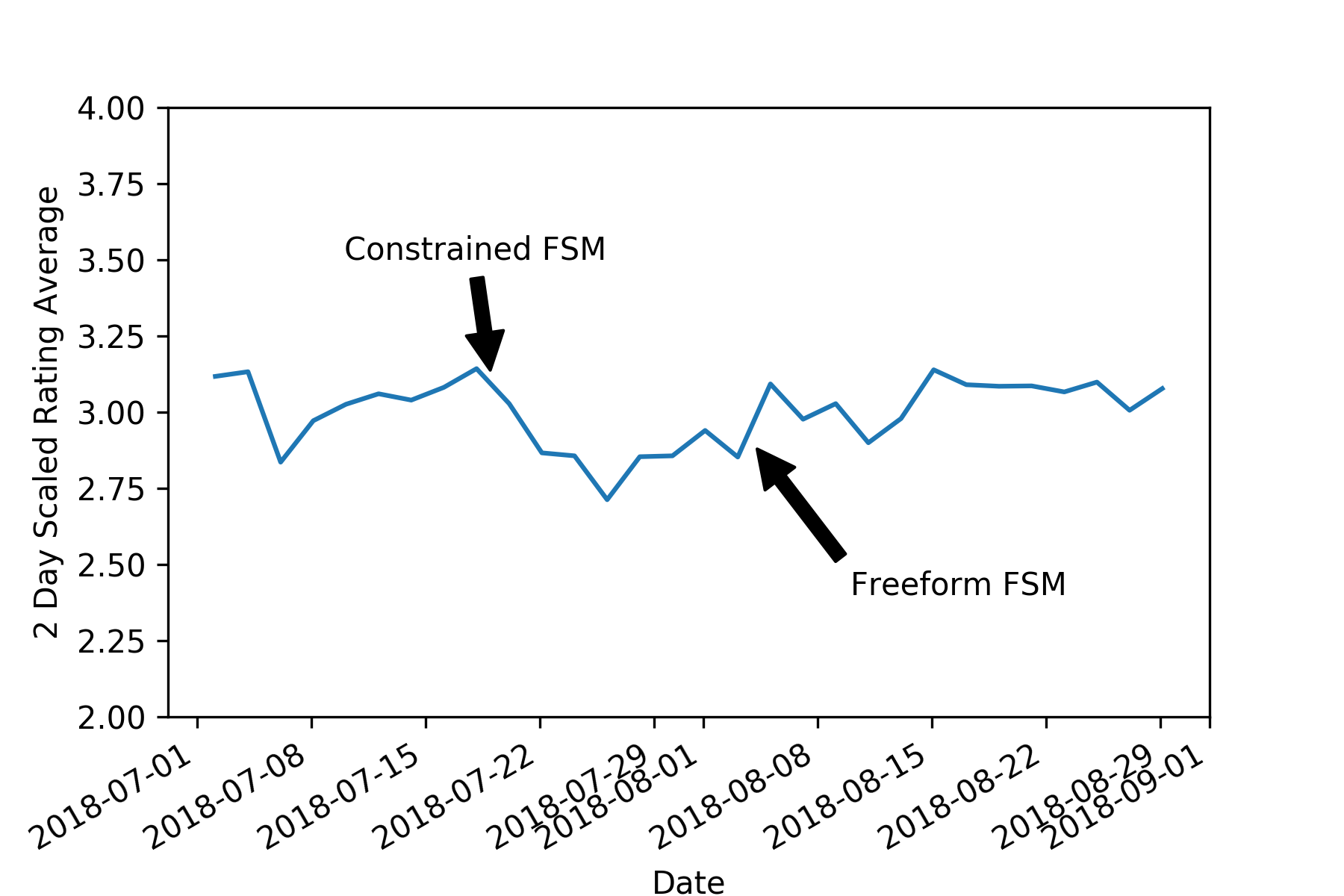}
\caption{Average scaled conversation rating vs date. The ratings have been scaled to have a mean value of 3.}
\label{figure:ratings}
\end{figure}

During the competition we also experimented with a more constrained Base FSM which always attempted to steer the user towards one of our Topic FSMs, and would loop back to a different Topic FSM when the user finished the previous one. As shown in Figure \ref{figure:ratings}, our ratings immediately took a severe hit. This result was surprising, as our hypothesis had been that the more we stay inside an FSM and not venture into "unknown" response generator territory, the better our conversations would be. Unfortunately, this was not the case. We observed that users were not looking for a limited set of perfect FSMs to explore. Rather, they preferred a higher level of novelty and spontaneity. The next, and current, iteration of our Base FSM, the Freeform FSM, avoided this mistake by balancing steering the user towards an FSM, asking exploratory questions (e.g. what is your favorite color?), attempting small talk, or allowing the user to propose a topic of conversation and then attempting to respond coherently. In Figure \ref{figure:ratings} the introduction of our Freeform FSM, which weakened the FSM Manager's bias towards Topic FSMs, is labeled and one can see an immediate uptick in the average conversation rating. This analysis suggests that despite the clear overall benefit of adding FSMs to our bot, it is important to understand that how changes are implemented is often just as important as what changes are made. Although FSMs are, in our opinion, essential to any real-world conversational bot, the FSM implementation must be done with care and utilize insights gained from analyzing human-human and human-agent conversations in order to conform to the conversational rules and standards of the human mind.
\subsection{Conversation Ratings}
To evaluate conversation quality, we developed a model to predict conversation ratings using actual conversations from the Alexa Prize competition. We train on 15,576 conversations, which are manually rated by the interlocutor. We use an LSTM with an embedding layer, Adam optimizer and mean squared loss for this task and evaluated with a 4-fold cross validation. We tokenize utterances and replace low frequency words. Our best model gives an RMSE of 1.04 averaging across all folds.

Using this model, we conducted an experiment to evaluate whether the first or the last half of the conversation has a higher impact on the rating model. When we evaluated using the first and second halves of the conversations, the RMSE was 1.44 and 1.24 respectively. This suggests that the last impression with Alexa has more impact that than the first. But, we still need the whole conversation for the best rating model. We note that this analysis is a first step at exploring the impact of early vs late stage conversation quality on the interlocutor's evaluation. There are a number of possible conflating factors that makes it difficult to directly analyze the halves of a conversation, such as the first half of the conversation being more structured and the fact that, by definition, conversation breakdowns that result in the user ending a session must occur at the end of the conversation.

\section{Scientific Contributions and Future Work}
Our goal for this year's Alexa Prize competition was to make progress on what we believe to be key challenges for socialbots. Our previous experience has shown us that coherent conversation is difficult to create using corpus-based techniques that confine themselves to a limited context. At the same time, the use of handcrafted and scripted conversations, while capable of generating reasonable dialogs, does not appear to get the essence of good discourse, which we believe requires equal participation by both interlocutors. Such constrained interactions tend to shift control to the socialbot, and while this can produce extended interaction, we do not believe it results in a satisfying user experience.

We consequently chose to focus on the development of a baseline conversational intelligence. We did this in two ways. The first was creating a semantic grammar that would allow the bot to understand the mechanics of an ongoing exchange. We believe that we were successful in achieving this goal, though much work remains to be done. About three-quarters of otherwise unlabeled concepts were identified and utilized to create the above mentioned grammar. This conversational grammar is a beginning and would need to be rationalized and further developed.

The other challenge in creating conversations is maintaining cohesion and context management. Functions such as co-reference and topic management, perhaps subsumed under a well-organized history function are essential. Our approach was to first develop a significantly flexible FSM capability, and to cast most of what went on in the system in terms of FSMs, even for very simple actions such as responding to \texttt{Backstory} questions. This, as well as the ability to easily manage sets of FSMs through mechanisms such as a stack and the use of null arcs, was designed to create a simple authoring environment. The further goal would be to develop an approach to induce FSMs from a collected conversation corpus of successful exchanges. In principle, this would allow a socialbot to improve its conversational skills, as well as identify topics that appear to engage users.

\section{Conclusion}
Engaging socialbots are difficult to build, in part because there is no clear scientific foundation for computational conversation. We believe that over the course of the Alexa Prize, we and others have begun to identify key elements that need to be created. Specifically, we identified context tracking and dialog management to be core weaknesses in conversational artificial intelligence.  We have contributed a framework for dynamic Finite State Machines to control and maintain the flow of conversations, without constraining the responses or topics about which our socialbot can converse. Our dialog management, combined with a mixture of scripted, partially scripted, and retrieved responses allowed our bot to reach a peak rank of 4th place in the Alexa Prize competition.

\section{Acknowledgments}
We wish to thank Pravalika Avvaru and Poorva Rane for their help engineering the Tartan socialbot. We also thank Zarana Parekh, Sreyashi Nag, Soham Ghosh, Anirudha Rayasam, Aditya Siddhant, and Radhika Parik for their help developing Tartan as part of a course project. Lastly, we thank Alan Black, Jamie Callan, and Eduard Hovy for their helpful discussion and feedback.

\bibliography{alexaprize_2017,tartan_proposal,ap_2018}

\end{document}